\pdfoutput=1
\documentclass[journal]{IEEEtran}
\usepackage{amsmath,amsfonts}
\usepackage{algorithmic}
\usepackage{algorithm}
\usepackage{array}
\usepackage[caption=false,font=normalsize,labelfont=sf,textfont=sf]{subfig}
\usepackage{textcomp}
\usepackage{stfloats}
\usepackage{url}
\usepackage{verbatim}
\usepackage{graphicx}
\usepackage{cite}
\usepackage{amsmath}
\usepackage{multirow}
\usepackage{booktabs}
\usepackage{caption}
\usepackage{amssymb}
\begin{document}

\title{Relevance-guided Audio Visual Fusion for \\ Video Saliency Prediction}

\author{
Li Yu, 
Xuanzhe Sun,
Pan Gao,
Moncef Gabbouj,
\thanks{Li Yu is with the School of Computer Science, Nanjing University of Information Science and Technology, Nanjing 210044, China, and also with the Jiangsu Collaborative Innovation Center of Atmospheric Environment and Equipment Technology, Nanjing University of Information Science and Technology, Nanjing 210044, China. (e-mail: mailofyuli@126.com)}
\thanks{Xuanzhe Sun is with the School of Computer science, Nanjing University of Information Science and Technology, Nanjing 210044, China.}
\thanks{Pan Gao is with the College of Computer and Technology, Nanjing University of Aeronautics and Astronautics, Nanjing 211106, China}
\thanks{Moncef Gabbouj is with the Faculty of Information Technology and Communication Sciences,Tampere University,33101 Tampere, Finland}
}

\markboth{Journal of \LaTeX\ Class Files,~Vol.~14, No.~8, August~2021}%
{Shell \MakeLowercase{\textit{et al.}}: A Sample Article Using IEEEtran.cls for IEEE Journals}

\IEEEpubid{0000--0000/00\$00.00~\copyright~2021 IEEE}

\maketitle

\begin{abstract}
Audio data, often synchronized with video frames, plays a crucial role in guiding the audience’s visual attention. Incorporating audio information into video saliency prediction tasks can enhance the prediction of human visual behavior. However, existing audio-visual saliency prediction methods often directly fuse audio and visual features, which ignore the possibility of inconsistency between the two modalities, such as when the audio serves as background music. To address this issue, we propose a novel relevance-guided audio-visual saliency prediction network dubbed AVRSP. Specifically, the Relevance-guided Audio-Visual feature Fusion module (RAVF) dynamically adjusts the retention of audio features based on the semantic relevance between audio and visual elements, thereby refining the integration process with visual features. Furthermore, the Multi-scale feature Synergy (MS) module integrates visual features from different encoding stages, enhancing the network’s ability to represent objects at various scales. The Multi-scale Regulator Gate (MRG) could transfer crucial fusion information to visual features, thus optimizing the utilization of multi-scale visual features. Extensive experiments on six audio-visual eye movement datasets have demonstrated that our AVRSP network achieves competitive performance in audio-visual saliency prediction. 
\end{abstract}


\section{Introduction}
\IEEEPARstart{V}{ideo} saliency analysis encompasses two primary branches: Video Salient Object Detection (VSOD) and Video Fixation point Prediction (VFP). VSOD focuses on detecting and emphasizing objects that quickly capture viewers' attention because of their distinct color, brightness, shape, or motion, making them stand out from the background. In contrast, VFP seeks to predict the focal points of viewers' attention during video viewing, thereby emulating human visual and cognitive functions. The significance of VFP not only aids in understanding human visual cognitive processes but also offers critical insights for downstream tasks like region-of-interest-based video coding and video streaming. Accurately predicting viewers’ gaze enhances the optimization of video delivery and presentation, enriches the user experience, and broadens opportunities for video content creation, editing, and interactive design. Given the importance of VFP, this paper will explores this task in depth.

\begin{figure}[!t]
\centering
\includegraphics[width=3.6in]{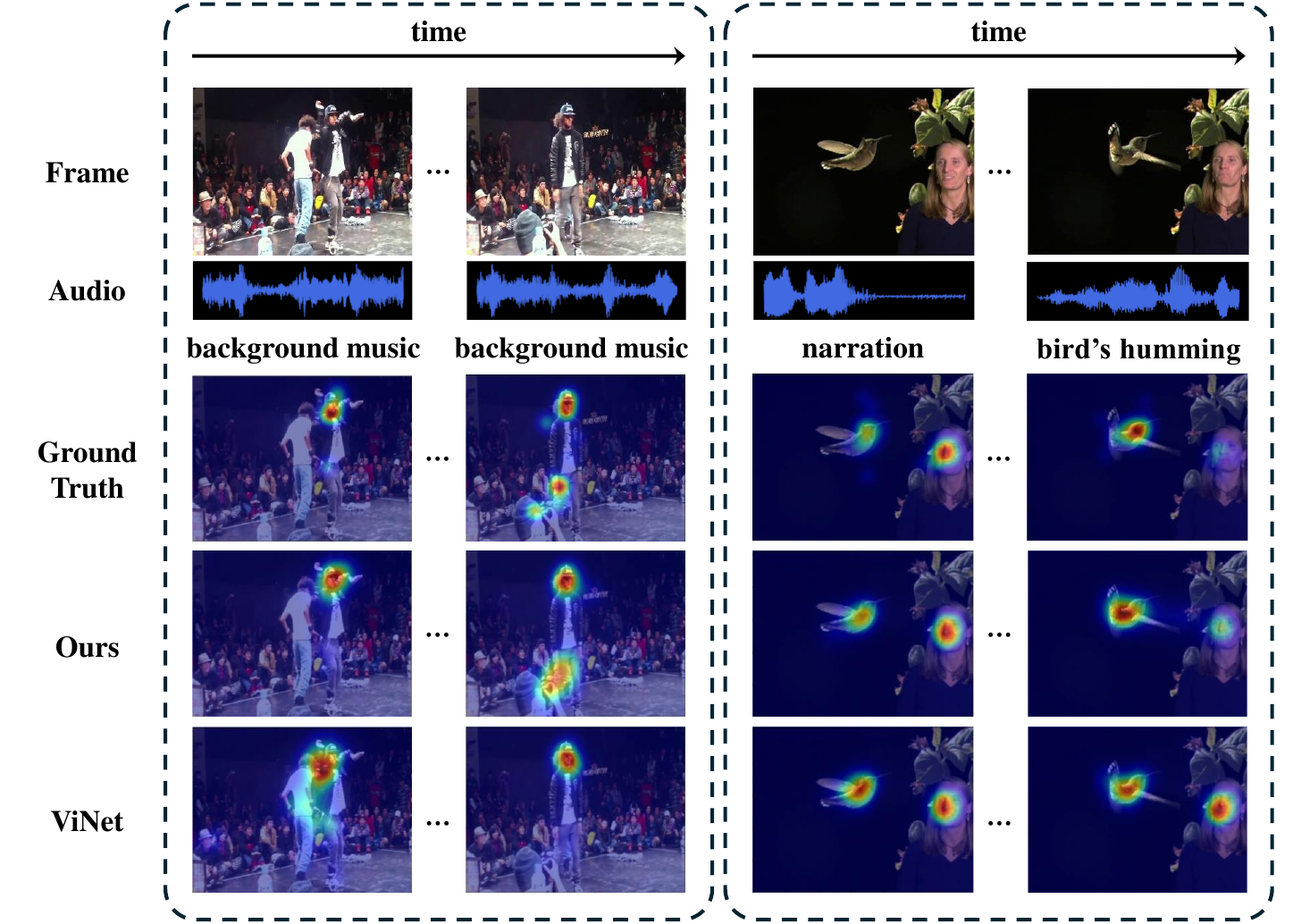}
\caption{The visualization results of the saliency prediction in a multimodal setting. For the video sequence on the left, which includes background music, our model minimizes the influence of irrelevant audio and focuses on key visual elements. For the video sequence on the right, the former half features a narration by a woman, during which our model prioritizes her presence. In the latter half, where only bird humming is present, our model shifts its attention to the bird.}
\label{fig:av relevance example}
\end{figure}

\IEEEpubidadjcol

Considering the inherent synchronization of audio with video content, several research  \cite{tsiami2020stavis, jain2021vinet, chen2021audiovisual, xiong2023casp} have focused on incorporating audio data into video saliency prediction frameworks. Tsiami \emph{et al.}\cite{tsiami2020stavis} created a spatio-temporal audio-visual saliency network using a unified architecture that integrates visual and auditory information in multiple stages. Similarly, Jain \emph{et al.} \cite{jain2021vinet} developed a methodology using a fully convolutional encoder-decoder architecture, which incorporates a bilinear fusion strategy to merge visual and auditory features. However, these approaches may fail to address the issue of semantic relevance when audio and visual data are synchronized in time but perform different roles, such as when the audio serves only as background music. Such an indiscriminate fusion of audio-visual features can reduce the predictive accuracy of these models. Xiong \emph{et al.} \cite{xiong2023casp} proposed a consistency-aware audio-visual saliency prediction network (CASP-Net), which aims to address temporal inconsistencies between audio and visual streams by employing a consistency-aware predictive coding module. While this method improves the alignment between audio and visual features, it still has limitations. The reliance on iterative consistency correction can be computationally intensive and may not fully exploit the dynamic relevance between modalities in complex scenes. To overcome these limitations, our approach employs a Relevance-guided Audio-Visual feature Fusion module (RAVF) that adaptively adjusts the integration level of audio features based on the semantic relevance between visual and audio features. This method ensures that only the relevant audio features are integrated with the visual features.

Ensuring semantic alignment between audio and visual elements is crucial in video content analysis due to the diversity of sound sources, such as dialogues and background music. By thoroughly exploring the latent semantic correlations of cross-modal signals, the proposed AVRSP method can rectify potential inconsistencies between different modalities. Specifically, we propose a Relevance-guided Audio-Visual Fusion method (RAVF), which calculates the semantic correlation between audio and visual features through cross-attention and efficiently utilizes relevant audio. To further enhance the model’s ability to detect salient objects of varying sizes, a Multi-scale feature Synergy module (MS) and Multi-scale Regulator Gate unit (MRG) have been designed. The main contributions of this work are summarized as follows:
\begin{itemize}
\item A Relevance-guided Audio-Visual feature Fusion module (RAVF) adaptively modulates the incorporation of audio features based on their semantic relationships with visual content.

\item A Multi-scale feature Synergy (MS) module aggregates visual information from various encoding layers to enhance the model's representation capability in predicting salient objects across different scales.

\item Through the Multi-scale Regulator Gate (MRG), the network could channels essential fusion information to visual components, refining the saliency prediction process by making better use of the multi-scale visual data.

\end{itemize}

\section{Related works}
\subsection{Visual Only Saliency Prediction}
With the public availability of large dynamically salient video datasets such as Hollywood-2\cite{marszalek2009actions} and DHF1K\cite{wang2018revisiting}, deep learning-based video saliency prediction methods\cite{ma2022video, gorji2018going, cornia2018predicting, zhou2021apnet, lou2022transalnet, huang2015salicon, pan2016shallow, kruthiventi2017deepfix, cong2022psnet} have flourished. Considering that video data contains both temporal and spatial dimensions, current research\cite{wang2018revisiting, linardos2019simple,chen2021video} primarily adopts a framework combining Convolutional Neural Networks (CNNs) and recursive models. These methods typically use CNNs to extract spatial features, and to sense temporal information, the spatial features of each frame are fed into the input gate of their respective recursive models separately. Finally, a decoder is applied to generate saliency predictions. Wang \emph{et al.} \cite{wang2018revisiting} proposed a framework that combines CNN-LSTM and the attention mechanism, integrating static and dynamic saliency information. The framework enhances the capture of static features through the attention mechanism, while using LSTM to handle dynamic changes between video frames. Linardos \emph{et al.} \cite{linardos2019simple} proposed integrating an LSTM module in the intermediate stage of a CNN-based encoder-decoder structure to model the outputs of different stage encoders, then feeding the represented spatio-temporal information into the decoder to obtain the predicted saliency map. Compared to traditional methods that process one frame at a time with LSTM, Chen \emph{et al.} \cite{chen2021video}, aiming to enhance the network's perception of temporal information, chose to input 3 frames of data into the network simultaneously, further enhancing the network's ability to capture temporal information. Moreover, considering that spatial displacement between consecutive frames might cause feature misalignment, affecting the learning process and the clarity of prediction results, they also used deformable convolution techniques to preprocess features, ensuring the features are aligned before entering LSTM. 

3D convolution, with its ability to simultaneously perceive temporal and spatial information, has also recently been explored by researchers for this task\cite{yao2021deep,min2019tased,bellitto2021hierarchical,bak2017spatio}. Min \emph{et al.} \cite{min2019tased} introduced a 3D fully convolutional encoder-decoder network, TASED-Net, focused on the aggregation of spatial and temporal features. In this network, the encoder encodes input frames into low-resolution spatio-temporal features, and the prediction network decodes these features, aggregating temporal information to generate full-resolution saliency maps. Bellitto \emph{et al.} \cite{bellitto2021hierarchical} proposed a multi-branch encoder-decoder network. This model introduces a saliency network and domain adaptation mechanism that utilizes different scales of spatio-temporal feature extraction, with each scale branch predicting a saliency map at a specific level of abstraction. These saliency maps are then combined to produce the final prediction result. Bak \emph{et al.} \cite{bak2017spatio} proposed two types of single-stream convolutional neural networks that process spatial and temporal information separately and employ different strategies such as direct averaging, maximum fusion, and convolutional fusion to integrate appearance and motion features.

\subsection{Audio-Visual Saliency Prediction}
Audio often appears alongside video and can impact human visual fixation points, demonstrating its ability to guide or change people's visual focus. When viewers receive specific audio signals, such as environmental sounds or dialogue, their attention is naturally drawn to the visual elements related to the sound. In addition, sound can enhance the perception of a scene's context, helping viewers to build a richer and more concrete understanding of visual content. Simultaneously, sound can affect people's emotions and expectations, thereby indirectly adjusting their visual focus.

Although sound has a significant impact on saliency, research on deep learning-based audio-visual saliency attention prediction is still in its infancy, with only a few works dedicated to this area. Tavakoli \emph{et al.} \cite{tavakoli2019dave} adopted a dual-stream neural network architecture, using a 3D residual network to process audio and video input separately. The video branch processes image frames, while the audio stream branch processes log Mel spectrogram frames of the audio signal. The model merges these processed data streams and generates the final saliency map through a series of upsampling and convolutional layers. Tsiami \emph{et al.} \cite{tsiami2020stavis} integrated visual features extracted by SUSiNet \cite{koutras2019susinet} and auditory features extracted by SoundNet \cite{aytar2016soundnet}, achieving precise localization of sound sources through a single network. Chen \emph{et al.} \cite{chen2021audiovisual} introduced a multisensory framework for video saliency prediction that integrates audio and visual signals using a deep learning architecture. The model consists of four modules: auditory feature extraction with 1D CNNs, visual feature extraction with VGG16, semantic interaction, and feature fusion. By combining these extracted features, the model generates the final saliency map. Jain \emph{et al.} \cite{jain2021vinet} used a 3D convolution-based encoder-decoder architecture, with its backbone network employing the S3D \cite{xie2018rethinking} network pre-trained on the Kinetics dataset to encode video segments. The S3D network includes 3D convolutional layers that can effectively encode spatio-temporal information and contains multiple convolutional blocks to extract features at different scales. The decoder introduces skip connections from the encoder and uses 3D convolutions and upsampling layers for saliency prediction. Xiong \emph{et al.} \cite{xiong2023casp}, based on theoretical neuroscience, proposed a predictive coding module with consistency perception capability to iteratively improve the consistency between audio and video representations.

The recent work of Zhu \emph{et al.} \cite{zhu2024discrete} introduces an implicit neural representation-based model (INR), DAVS, which maps spatial-temporal coordinates to corresponding saliency values, effectively learning compact feature representations and incorporating continuous dynamics of videos in audio-visual saliency prediction. This model leverages a parametric neural network for adaptive feature fusion, capturing intrinsic interactions across modalities and self-adaptively integrating audio and visual cues. Another noteworthy approach is the MTCAM model\cite{zhu2024mtcam}, which employs a multi-modal transformer-based class activation mapping technique to convert video category labels to pseudo-fixations in a weakly-supervised manner. This model demonstrates significant improvements in audio-visual saliency prediction by leveraging cross-modal transformers and efficient feature reuse mechanisms.

A significant challenge in audio-visual saliency prediction is ensuring the relevance of fused features. Many existing models directly combine audio and visual data without considering their contextual relationship, which can lead to reduced accuracy. Our approach addresses these limitations through the introduction of the Relevance-Guided Audio-Visual Feature Fusion (RAVF) module. which can dynamically adjusts the integration of audio features based on their semantic relevance to the visual content. This relevance-guided fusion significantly improves the model's ability to handle various scenarios.

\section{Methodology}
\begin{figure*}[!t]
\centering
\includegraphics[width=7in]{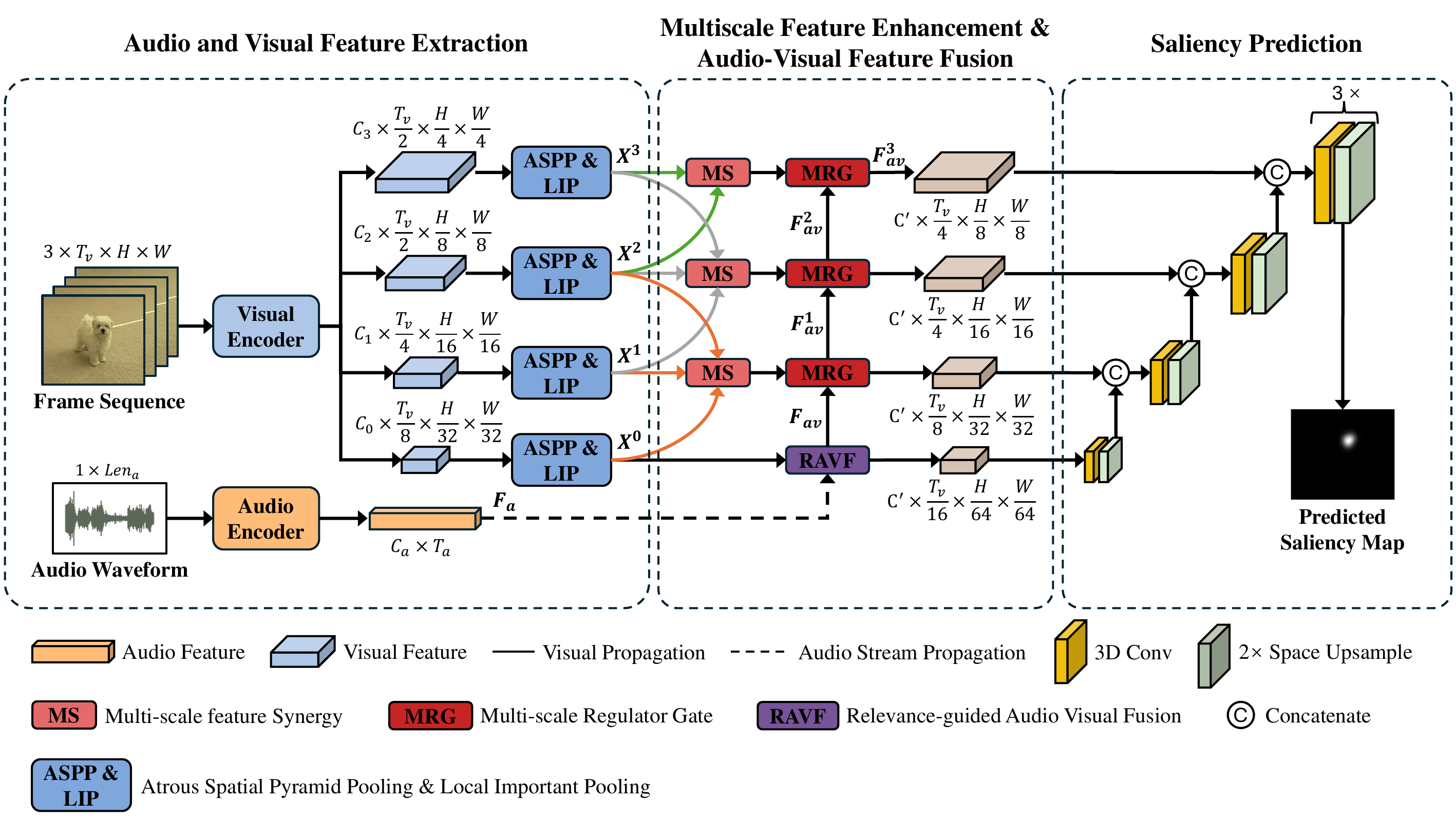}
\caption{
The proposed AVRSP, mainly consists of three main stages: (1) Audio and Visual Feature Extraction, where audio waveforms and frame sequences are encoded using SoundNet and S3D models respectively, (2) Multi-scale Feature Enhancement \& Audio-Visual Feature Fusion, where extracted features undergo dynamic fusion through the Relevance-guided Audio-Visual Fusion (RAVF) module. The Multi-Scale feature Synergy (MS) module along with the Multi-scale Regulator Gate (MRG) adjust and enhance feature interplay, (3) Saliency Prediction, and several saliency decoder blocks are used to estimate the saliency map from the multi-scale audio-visual features.
}
\label{fig:overview}
\end{figure*}

In order to improve the accuracy of video saliency prediction and address the often-overlooked issue of semantic inconsistency between audio and visual features, we propose a novel relevance-guided audio-visual saliency prediction network, named AVRSP. The overall framework is shown in Figure \ref{fig:overview}. The AVRSP comprises three main stages: (1) Audio and Visual Feature Extraction, where the corresponding features are extracted from audio waveforms and video frames; (2) Multi-scale Feature Enhancement and Audio-Visual Feature Fusion, which effectively enhance and fuse these features at different scales; and (3) Saliency Prediction, utilizing the fused features of different scales to predict the final saliency maps. At the core of the AVRSP network is the Relevance-guided Audio-Visual feature Fusion module (RAVF), which adaptively adjusts the degree of fusion based on the semantic relevance between audio and visual features. Additionally, the AVRSP includes a Multi-Scale feature Synergy module (MS) to enhance its ability to represent objects of different sizes, and a Multi-Scale Regulator Gate (MRG) to transfer crucial information derived from the audio-visual fusion process into the multi-scale visual representations.
These modules collectively form a crucial part of the Multi-scale Feature Enhancement \& Audio-Visual Feature Fusion stage. The subsequent sections of this paper will introduce these core components in detail.

\subsection{Backbone}
\textbf{Visual Backbone.} 
To simultaneously extract the temporal and spatial features within the video data, the proposed method utilizes a 3D convolution, the S3D\cite{xie2018rethinking}, as the visual backbone.

Specifically, consecutive video frames are input into S3D, where four encoder blocks output multi-scale features with different receptive fields. To further enrich the representation ability of the model, Atrous Spatial Pyramid Pooling (ASPP) \cite{chen2017deeplab} and Local Important Pooling (LIP) \cite{gao2019lip} are introduced after each encoder block, resulting in multi-scale features $X^{i}$ (i=0,1,2,3) for subsequent processing.

\textbf{Audio Backbone.} For the audio stream, we preserve the original audio data structure as much as possible, similar to \cite{tsiami2020stavis}. Meanwhile, the corresponding audio segments are captured according to the video frames. At the same time, a Hanning window is added to alleviate the edge effect caused by segmentation. For feature extraction, we selected SoundNet \cite{aytar2016soundnet} as our audio backbone based on its superior performance in learning audio features, especially its ability to handle multimodal data. By learning from a large number of unlabeled videos via a teacher-student network, SoundNet captures the intrinsic correlations between complex audio and visual patterns, which is crucial for our task. The extracted features, denoted as $F_a$, are then combined with visual features in our fusion module for further processing.

\subsection{Relevance-Guided Audio-Visual feature Fusion}

\begin{figure*}[!t]
\centering
\includegraphics[width=6in]{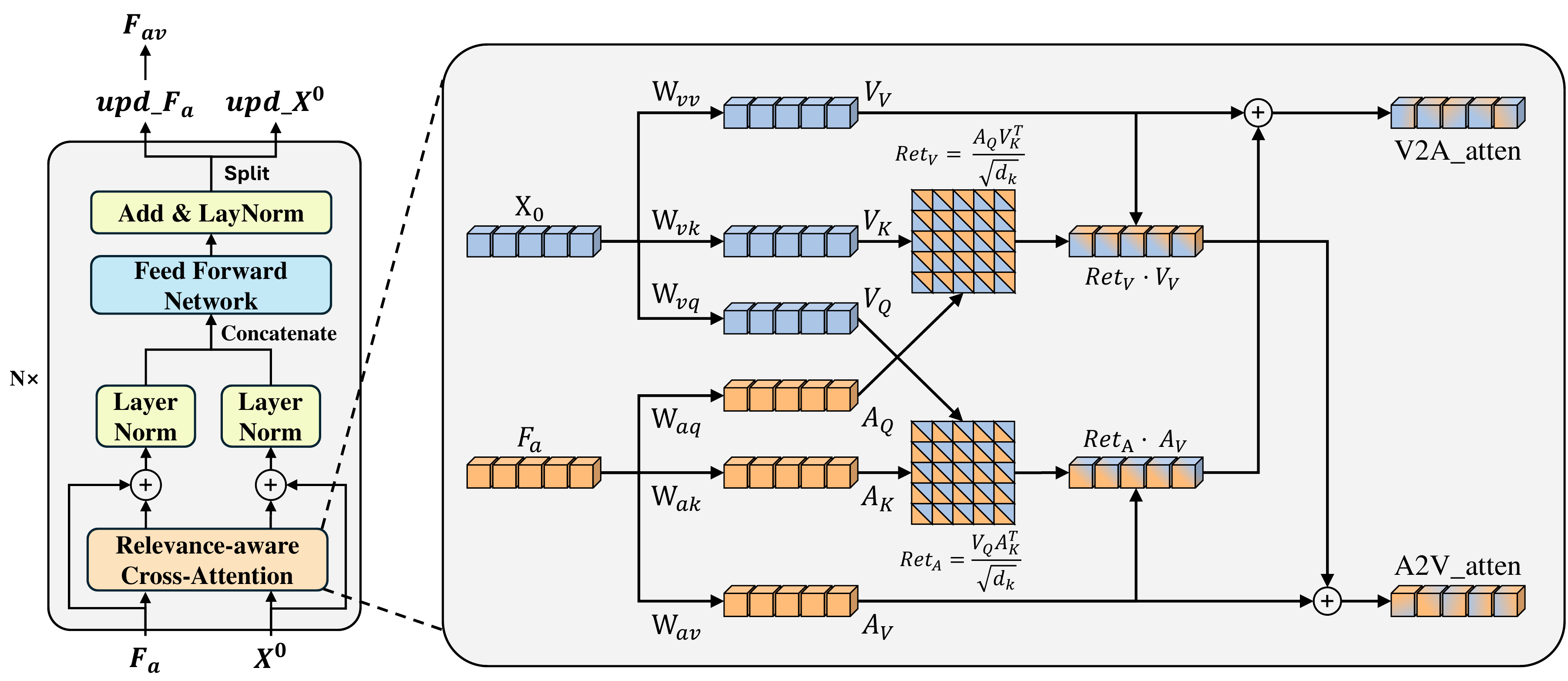}
\caption{The illustration of the Relevance-Guided Audio-Visual Fusion (RAVF) method.}
\label{fig:RAVF}
\end{figure*}

The relevance between audio and visual data is crucial for effective multimodal feature fusion. To leverage this relevance, we propose a novel attention-based fusion method called Relevance-guided Audio-Visual Feature Fusion (RAVF), illustrated in Figure \ref{fig:RAVF}. In the RAVF block, the audio feature $F_{a}$ and the visual feature $X^0$ are first processed by the Relevance-aware Multihead Cross-attention layer, followed by a Residual connection and the Layer Normalization. The output then passes through a feedforward layer and another residual connection and Layer Normalization, iteratively refining the fused features through \emph{N} RAVF blocks.

\textbf{Relevance-aware Multihead Cross-attention}. This mechanism computes the relevance between audio and visual features by generating query, key, and value vectors for both modalities. Specifically, six different affine transformations ($W_{vq}, W_{vk}, W_{vv}, W_{aq}, W_{ak}$ and $W_{av}$) are applied to the visual feature $X^0$ and the audio feature $F_a$ to generate visual-query $V_Q$, visual-key $V_K$, visual-value $V_V$, audio-query $A_Q$, audio-key $A_K$, and audio-value $A_V$, respectively. The introduction of RetA and RetV represents the retention levels of the audio-value $A_V$ and visual-value $V_V$ , respectively. These metrics provide a measure of how much of each modality's features are retained. Unlike traditional self-attention mechanisms, this approach omits the use of softmax normalization, as RetA and RetV directly indicate the extent of feature retention, thus simplifying the process and directly reflecting the contribution of each modality to the final fused representation.

The visual-to-audio attention $v2a\_attn$ and audio-to-visual $a2v\_attn$ can be formulated as follows:
\begin{equation}
    V_i = f_{vi}(X^0), \quad A_i = f_{ai}(F_a), \quad i=q,k,v
\end{equation}
\begin{equation}
    Ret_A = \frac{V_Q A_K^T}{\sqrt{d_k}} 
\end{equation}
\begin{equation}
    Ret_V = \frac{A_Q V_K^T}{\sqrt{d_k}} 
\end{equation}
\begin{equation}
    \text{v2a\_attn} = Ret_A \cdot A_V + V_V
\end{equation}
\begin{equation}
    \text{a2v\_attn} = Ret_V \cdot V_V + A_V
\end{equation}

In the calculation of $v2a\_attn$, the attention scores are computed by querying visual features against audio keys, which determine the relevance from the video’s perspective. To better reflect this perspective, the visual values $V_V$ are added to the weighted audio features. Similarly, in $a2v\_attn$, the audio values $A_V$ are also added to the weighted visual features to enhance the integration from the audio’s perspective.

To further refine our multimodal fusion strategy, we incorporated a set of adaptive attention weights for each attention head. The module evaluates the integrated features of both modalities to produce weights that dynamically adjust the influence of each head, thus the fusion feature $F_{av}$ can be formulated as follows:
\begin{equation}
C_{weights} = \text{Softmax} \left( \text{MLP} \left( \text{Cat}(v2a\_attn, a2v\_attn) \right) \right)
\end{equation}
\begin{equation}
F_{av} = C_{weights} \cdot v2a\_attn
\end{equation}
The calculated weights $C_{weights}$ are then applied to modify the impact of each head, enhancing the relevance-guided aspect of the fusion. This method allows each head to focus adaptively on the most relevant features based on the correlation data, significantly improving the efficacy of audio-visual integration. This results in a multimodal feature fusion process that is more adaptive and contextually aware, adept at managing the complexities associated with varied multimodal inputs.

\subsection{Multi-scale Synergy and Multi-scale Regulator Gate}
\textbf{Multi-scale feature Synergy.}
\begin{figure}[!t]
\centering
\includegraphics[width=3.5in]{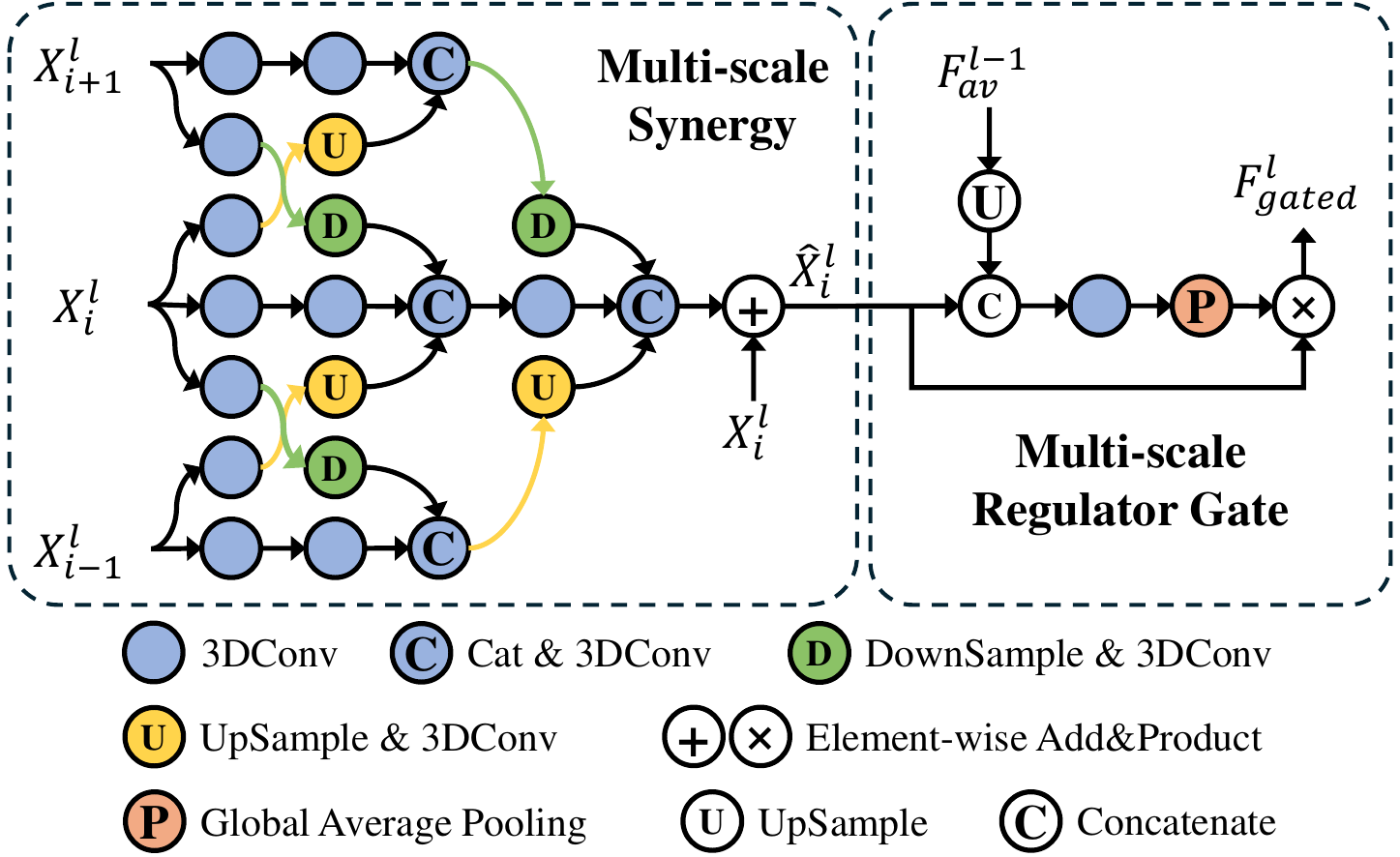}
\caption{Multi-scale feature Synergy and Multi-scale Regulator Gate.}
\label{fig:MS_MRG}
\end{figure}
In saliency prediction tasks, large objects are generally perceived as more salient, while objects that may be small in size but move rapidly or have a strong color contrast with their surroundings are also salient. These multi-scale features are often obtained through deep or shallow networks with different receptive fields. To accurately capture objects with these different characteristics, the proposed method introduces a Multi-Scale feature Synergy module (MS), whose structure is shown in the left half of Figure \ref{fig:MS_MRG}. The multi-scale feature $X^i, i=1,2,3$  interacts with adjacent features on various scales, which can be represented as:
\begin{equation}
\hat{X}^{m}=MS(X^{h},X^{m},X^{l})
\end{equation}
where $X^{h}$, $X^{m}$, $X^l$ represent the adjacent multi-scale feature maps from high resolution to low, respectively; $\hat{X}^{m}$ is the enhanced feature, and the MS mapping can be formulated as follows:
\begin{align}
X_1^i &= \mathcal{C}(X^i) && \text{for } i \in {h, m, l} \\
X_2^i &= \mathcal{U}(X_{1}^{i\_low}) + \mathcal{C}(X_{1}^{i}) + \mathcal{D}(X_{1}^{i\_high}) && \text{for } i \in {h, m, l}\\
X_3^m &= \mathcal{D}(X_2^h) + \mathcal{C}(X_2^m) + \mathcal{U}(X_2^l) \\
\hat{X}^{m} &= X_3^m + X^m
\end{align}
Among them, $\mathcal{C}$ represents a 3D convolution, $\mathcal{U}$ represents upsampling followed by 3D convolution, and $\mathcal{D}$ represents downsampling followed by 3D convolution. After processing through the MS module, the enhanced feature $(\hat{X}^{m})$  can be obtained.

\textbf{Multi-scale Regulator Gate.}
Traditionally, only fused features serve as the primary input to the decoder, overlooking the rich visual information present among different visual encoder blocks. To enhance our model’s ability to leverage this information, we incorporated a gating mechanism that selectively processes the lowest-resolution audio-visual features alongside higher-resolution visual features. This approach helps to optimize the utilization of multi-scale visual features.

Assume there are visual features $\hat{X}^{l}  (l=1,2,3)$ of different scales from the Multi-scale Synergy and the fused audio-visual feature $F_{av}^{l-1}$ from the RAVF block or the previous gate unit.
\begin{equation}
    \hat{F}_{av}^{l} = UpSample(F_{av}^{l-1})
\end{equation}
\begin{equation}
    G^{l}=GAP(\sigma(Conv(Cat(\hat{F}_{av}^{l}, \hat{X}^{l}))))
\end{equation}
\begin{equation}
    F_{gated}^{l}=Conv(G^{l} \odot \hat{F}_{av}^{l})
\end{equation}
where $Cat$ denotes the concatenate operation, $Conv$ denotes the 3D convolution layer, $\sigma$ denotes the sigmoid function, and $GAP$ denotes the global average pooling. The gate values $G^{l}$ are then applied for weighting the audio-visual features $F_{av}^{l}$.

\subsection{Saliency Generation}
The fused audio-visual feature $F_{av}$ is fed into the decoder synchronously with other multi-scale visual features $X_{3}^{'}$, $X_{2}^{'}$, and $X_{1}^{'}$ to predict the saliency map. Similarly to the encoder, the decoder contains four blocks, each consisting of a convolutional layer and an upsampling layer. The specific structure of the decoder is shown in the Saliency Prediction part of Figure \ref{fig:overview}.

\subsection{Loss Function}
We selected a combination of Kullback-Leibler divergence (KL), linear correlation coefficient (CC), and similarity measure (Sim) as the loss function to comprehensively evaluate the predicted saliency maps from different perspectives. Each metric contributes uniquely to the assessment of prediction errors, ensuring a well-rounded optimization of the model. 

First, Kullback-Leibler divergence (KL) measures the divergence between the predicted and ground truth saliency distributions. This metric captures the overall difference in probability distributions, quantifying how well the predicted map approximates the true distribution.    Given its ability to address the overall alignment of the saliency distribution, KL divergence is assigned the highest weight in the loss function.   The calculation for KL divergence is:
\begin{equation}
    \text{KL}(P,G)=\sum_{i=1}^{N} G_{i} \log \left(\frac{\varepsilon + G_{i}}{\varepsilon + P_{i}}\right)
\end{equation}
where $N$ is the total number of pixels in the image, and $\varepsilon$ is a tiny positive number to prevent division by zero. $P_i$ and $G_i$ are the normalized model prediction and the ground truth saliency map, respectively.

Second, the linear correlation coefficient (CC) evaluates the linear relationship between the predicted and ground truth saliency maps. This metric assesses how well the variations in the predicted map correspond to those in the ground truth map, indicating the strength and direction of their relationship. High correlation implies that the predicted saliency map accurately reflects the relative importance of different regions, ensuring that the model captures the underlying patterns in the data. The calculation for CC is:
\begin{equation}
    \text{CC}(P,G)=\frac{\sum_{i=1}^{N} (P_{i}-\bar{P})(G_{i}-\bar{G})}{\sqrt{\sum_{i=1}^{N} (P_{i}-\bar{P})^2} \sqrt{\sum_{i=1}^{N}(G_{i}-\bar{G})^2}}
\end{equation}
where $\bar{P}$ and $\bar{G}$ are the mean values of P and G, respectively. This formula computes the normalized cross covariance between P and G, which can be understood as the similarity of two saliency maps.

Lastly, the similarity measure (Sim) assesses the degree of overlap between the predicted and ground truth saliency maps. This metric highlights regions where both maps agree, providing a measure of their similarity in terms of saliency values. By maximizing this measure, the model improves the alignment of salient regions, enhancing the local accuracy of the predictions. The calculation for Sim is:
\begin{equation}
    \text{Sim}(P,G)=\sum_{i=1}^{N} \min(P_{i}, G_{i})
\end{equation}
where $P_{i}$ and $G_{i}$ are the normalized values at each pixel.

The total loss function is thus formulated as:
\begin{equation}
    \text{Loss}(P,G) = \text{KL}(P,G)+\alpha_1 \ast \text{CC}(P,G)+\alpha_2 \ast \text{Sim}(P,G)
\end{equation}
where $\alpha_1$ and $\alpha_2$ are hyperparameters, set to -0.1 and -0.1 respectively. By combining these three metrics, the loss function ensures that the model produces saliency maps that not only match the ground truth distribution but also maintain strong linear relationships and high local accuracy.

\section{Experiment}
\subsection{Dataset}
We conducted a comparative assessment against existing works on the commonly used visual-only dataset DHF1K\cite{wang2018revisiting} and six audio-visual datasets, including DIEM\cite{mital2011clustering}, Coutrot1\cite{coutrot2014saliency}, Coutrot2\cite{coutrot2016multimodal}, AVAD\cite{min2016fixation}, ETMD\cite{koutras2015perceptually}, and SumMe\cite{gygli2014creating}.

\textbf{DHF1K}: DHF1K consists of 1000 video sequences, which include 600 training videos, 100 validation videos, and 300 test videos. DHF1K encompasses video sequences of various themes, and we chose to pre-train the visual branch of our model on this dataset.

\textbf{Coutrot1}: Coutrot1 enriches the viewer's experience with 60 videos across four thematic categories: individual and group dynamics, natural vistas, and close-ups of faces, supported by visual attention data from 72 contributors.

\textbf{Coutrot2}: In contrast, Coutrot2 focuses on a more niche setting, capturing the interactions of four individuals in a conference setting, with eye-tracking data from 40 observers.

\textbf{DIEM}: The DIEM dataset is even more diverse, containing 84 video clips divided into 64 training and 17 test sets, covering fields such as commercials, documentaries, sports events, and movie trailers. Each video is accompanied by eye-tracking fixation annotations from approximately 50 viewers in a free-viewing mode.

\textbf{AVAD}: The AVAD dataset is a set of 45 brief video sequences, each lasting between 5 and 10 seconds, including a spectrum of dynamic audio-visual experiences, such as musical performances, sports activities, and journalistic interviews. This dataset is enhanced with eye-tracking insights collected from 16 individuals.

\textbf{ETMD}: The ETMD dataset draws its content from a selection of Hollywood cinematic productions, encapsulating 12 distinct film excerpts. This dataset is meticulously annotated with the visual tracking data of 10 evaluators.

\textbf{SumMe}: Completing the collection, the SumMe dataset offers a varied palette of 25 video vignettes, capturing everyday leisure and adventure activities ranging from sports to culinary arts and travel explorations. The visual engagement of viewers is quantified through eye-tracking data from 10 participants.

\begin{table*}[!t]
\renewcommand{\arraystretch}{1.1}
\caption{COMPARISON RESULT OF DIFFERENT METHODS ON THE DIEM, ETMD AND AVAD DATASETS}
\label{table:comparison with sota part1}
\centering
\scalebox{1.1}{
\begin{tabular}{l|cccc|cccc|cccc}
\hline
\multirow{2}{*}{Method} & \multicolumn{4}{c|}{DIEM}                                        & \multicolumn{4}{c|}{ETMD}                                          & \multicolumn{4}{c}{AVAD}                                         \\
                        & SIM$\uparrow$  & CC$\uparrow$   & NSS$\uparrow$ & AUC-J$\uparrow $ & SIM$\uparrow$ & CC$\uparrow$   & NSS$\uparrow$ & AUC-J$\uparrow$ & SIM$\uparrow$  & CC$\uparrow$   & NSS$\uparrow$ & AUC-J$\uparrow$\\ 
\hline
TASED-Net(V)\cite{min2019tased}       & 0.461          & 0.557          & 2.16          & 0.881          & 0.366           & 0.509          & 2.63          & 0.916           & 0.439          & 0.601          & 3.16          & 0.914          \\
STAViS(V)\cite{tsiami2020stavis}      & 0.472          & 0.567          & 2.19          & 0.879          & 0.412           & 0.560          & 2.84          & 0.929           & 0.443          & 0.604          & 3.07          & 0.915          \\
ViNet(V)\cite{jain2021vinet}          & 0.483          & 0.626          & 2.47          & 0.898          & 0.409           & 0.569          & 3.06          & 0.928           & 0.504          & 0.694          & \textbf{3.82} & 0.928          \\
CASP-Net(V)\cite{xiong2023casp}       & 0.538          & 0.649          & 2.59          & 0.904          & \textbf{0.471}  & \textbf{0.616} & 3.31          & \textbf{0.938}  & 0.526          & 0.681          & 3.75          & 0.931          \\
TSFP-Net(V)\cite{chang2021temporal}   & 0.527          & 0.651          & 2.62          & 0.906          & 0.433           & 0.576          & 3.09          & 0.932           & 0.530          & 0.688          & 3.79          & 0.931          \\
Ours(V)                               & \textbf{0.545} & \textbf{0.668} & \textbf{2.67} & \textbf{0.907} & 0.470           & 0.611          & \textbf{3.33} & 0.930           & \textbf{0.533} & \textbf{0.697} & 3.75          & \textbf{0.933} \\ 
\hline
STAViS(AV)\cite{tsiami2020stavis}     & 0.482          & 0.580          & 2.26          & 0.884          & 0.425           & 0.569          & 2.94          & 0.931           & 0.457          & 0.608          & 3.18          & 0.919          \\
ViNet(AV)\cite{jain2021vinet}         & 0.498          & 0.632          & 2.53          & 0.899          & 0.406           & 0.571          & 3.08          & 0.928           & 0.491          & 0.674          & 3.77          & 0.927          \\
CASP-Net(AV)\cite{xiong2023casp}      & 0.543          & 0.655          & 2.61          & 0.906          & \textbf{0.478}  & \textbf{0.620} & 3.34          & \textbf{0.940}  & 0.528          & 0.691          & 3.81          & \textbf{0.933} \\
TSFP-Net(AV)\cite{chang2021temporal}  & 0.527          & 0.651          & 2.62          & 0.906          & 0.428           & 0.576          & 3.07          & 0.932           & 0.521          & \textbf{0.704} & 3.77          & 0.932          \\
DAVS(AV)\cite{zhu2024discrete}        & 0.484          & 0.580          & 2.29          & 0.884          & 0.426           & 0.600          & 2.96          & 0.932           & 0.458          & 0.610          & 3.19          & 0.919          \\
Ours(AV)                              & \textbf{0.559} & \textbf{0.685} & \textbf{2.78} & \textbf{0.911} & 0.474           & 0.611          & \textbf{3.36} & 0.934           & \textbf{0.549} & \textbf{0.704} & \textbf{3.84} & \textbf{0.933} \\ 
\hline
\end{tabular}
}
\end{table*}

\begin{table*}[!t]
\renewcommand{\arraystretch}{1.1}
\caption{COMPARISON RESULT OF DIFFERENT METHODS ON THE COUTROT1, COUTROT2 AND SUMME DATASETS}
\label{table:comparison with sota part2}
\centering
\scalebox{1.1}{
\begin{tabular}{l|cccc|cccc|cccc}
\hline
\multirow{2}{*}{Method} & \multicolumn{4}{c|}{Coutrot1}                                    & \multicolumn{4}{c|}{Coutrot2}                                    & \multicolumn{4}{c}{SumMe}                                        \\
                        & SIM$\uparrow$  & CC$\uparrow$   & NSS$\uparrow$ & AUC-J$\uparrow $ & SIM$\uparrow$    & CC$\uparrow$     & NSS$\uparrow$   & AUC-J$\uparrow$  & SIM$\uparrow$    & CC$\uparrow$     & NSS$\uparrow$   & AUC-J$\uparrow$  \\ 
\hline
TASED-Net(V)\cite{min2019tased}      & 0.388          & 0.479          & 2.18          & 0.867          & 0.314          & 0.437          & 3.17          & 0.921          & 0.333          & 0.428          & 2.10          & 0.884          \\
STAViS(V)\cite{tsiami2020stavis}     & 0.384          & 0.459          & 1.99          & 0.862          & 0.447          & 0.653          & 4.19          & 0.941          & 0.332          & 0.418          & 1.98          & 0.884          \\
ViNet(V)\cite{jain2021vinet}         & 0.423          & 0.551          & 2.68          & 0.886          & 0.466          & 0.724          & 5.61          & 0.950          & 0.345          & 0.466          & 2.40          & 0.898          \\
CASP-Net(V)\cite{xiong2023casp}      & 0.445          & 0.559          & 2.64          & 0.888          & 0.567          & 0.756          & 6.07          & \textbf{0.963} & \textbf{0.382} & 0.485          & 2.52          & \textbf{0.904} \\
TSFP-Net(V)\cite{chang2021temporal}  & 0.447          & 0.571          & 2.73          & \textbf{0.895} & 0.528          & 0.743          & 5.31          & 0.959          & 0.362          & 0.463          & 2.28          & 0.894          \\
Ours(V)                              & \textbf{0.459} & \textbf{0.590} & \textbf{2.83} & \textbf{0.895} & \textbf{0.602} & \textbf{0.820} & \textbf{6.28} & \textbf{0.963} & 0.374          & \textbf{0.494} & \textbf{2.57} & 0.900          \\ 
\hline
STAViS(AV)\cite{tsiami2020stavis}    & 0.394          & 0.472          & 2.11          & 0.869          & 0.511          & 0.735          & 5.28          & 0.958          & 0.337          & 0.422          & 2.04          & 0.888          \\
ViNet(AV)\cite{jain2021vinet}        & 0.425          & 0.560          & 2.73          & 0.889          & 0.493          & 0.754          & 5.95          & 0.951          & 0.343          & 0.463          & 2.41          & 0.897          \\
CASP-Net(AV)\cite{xiong2023casp}     & 0.456          & 0.561          & 2.65          & 0.889          & 0.585          & 0.788          & 6.34          & 0.963          & \textbf{0.387} & 0.499          & 2.60          & \textbf{0.907} \\
TSFP-Net(AV)\cite{chang2021temporal} & 0.447          & 0.571          & 2.73          & 0.895          & 0.528          & 0.743          & 5.31          & 0.959          & 0.360          & 0.464          & 2.30          & 0.894          \\
DAVS(AV)\cite{zhu2024discrete}       & 0.400          & 0.482          & 2.19          & 0.869          & 0.512          & 0.734          & 4.98          & 0.960          & 0.339          & 0.423          & 2.29          & 0.889          \\
Ours(AV)                             & \textbf{0.463} & \textbf{0.595} & \textbf{2.87} & \textbf{0.899} & \textbf{0.617} & \textbf{0.837} & \textbf{6.47} & \textbf{0.964} & 0.381          & \textbf{0.504} & \textbf{2.72} & 0.903          \\ 
\hline
\end{tabular}
}
\end{table*}

\subsection{Implementation Details}
\textbf{Data Processing:} For the input of visual backbone, 32 consecutive video frames was randomly selected from one visual-only dataset and six audio-visual datasets, and each frame was scaled to 224$\times$384. To ensure temporal synchronization between the audio and video frames, we segmented the audio data based on the audio sampling rate and the video frame rate. Additionally, we applied a Hanning window during the segmentation process to reduce the impact of edge effects.

\textbf{Training Details:} The AdamW optimizer was utilized to optimize the proposed model, while a weight decay of 1e-5 was used for regularization. Additionally, we applied the learning rate scheduler ReduceLROnPlateau with a decay factor of 0.5 to automatically adjust the initial learning rate 1e-4 based on the validation loss. First, we train the visual branch for 80 epochs on DHF1K, where the feedforward process is as shown by the solid lines in Figure 1. After that, we combined it with the audio branch and trained it on six audio-visual datasets for 100 epochs. The input at this stage consists of 32 consecutive frames and their corresponding audio segments, the feedforward process is as shown by the dash lines in Figure 1.

\textbf{Testing Details:} For testing, we used a sliding window with a size of 32 and a step size of 1 to generate the corresponding saliency map for each frame. Since the model takes 32 consecutive frames as input and predicts the saliency map for the last frame, it cannot predict the saliency maps for the first 31 frames directly. To address this, we employed a reverse window strategy similar to \cite{jain2021vinet}. For example, to predict the saliency map of the first frame, we reversed the sequence from the 1st to the 32nd frame (i.e., the sequence becomes the 32nd frame, 31st frame, ..., 2nd frame, 1st frame). This reversed sequence is then used as the input to predict the saliency map for the first frame.

\subsection{Evaluation Metrics}
To compare with existing work, we selected four widely used evaluation metrics: CC, NSS, AUC-J, and SIM. CC is used to measure the linear correlation between the predicted saliency map and the ground truth saliency map. SIM measures the intersection distribution between the predicted saliency maps and the ground truth saliency maps, assessing the degree to which the two distributions match. AUC-J is used to compare the detected saliency map as a binary classifier with the true saliency map. NSS measures the average normalized saliency at the fixed positions of human eye fixations.

\subsection{Performance Comparison}
In this section, we compare our proposed network with recent saliency prediction methods on six different audiovisual datasets. The methods can be categorized into two groups: visual-only and audio-visual. The visual-only methods include TASED-Net(V)\cite{min2019tased}, STAViS (V)\cite{tsiami2020stavis}, ViNet (V)\cite{jain2021vinet}, CASP-Net (V)\cite{xiong2023casp}, and TSFP-Net (V)\cite{chang2021temporal}. These methods rely solely on visual data to predict saliency maps. The audio-visual methods incorporate both visual and audio data to enhance saliency prediction. The methods in this category include STAViS (AV), ViNet (AV), CASP-Net (AV), DAVS(AV)\cite{zhu2024discrete} and TSFP-Net (AV). The results are shown in Table \ref{table:comparison with sota part1} and \ref{table:comparison with sota part2}, the best performance is in bold. Specifically, Ours(V) and Ours(AV) achieve an average improvement of approximately 0.97\% and 1.58\% across all metrics on the six audio-visual datasets compared to CASP-Net (V) and CASP-Net (AV). Notably, our model performs well on datasets like DIEM and AVAD, which feature a wide range of dynamic and complex audio-visual content. The diversity and complexity of these datasets highlight the strengths of our relevance-guided fusion mechanism and multi-scale feature integration, enabling our model to adapt and excel in various challenging scenarios.

\begin{figure*}[ht]
\centering
\includegraphics[width=7in]{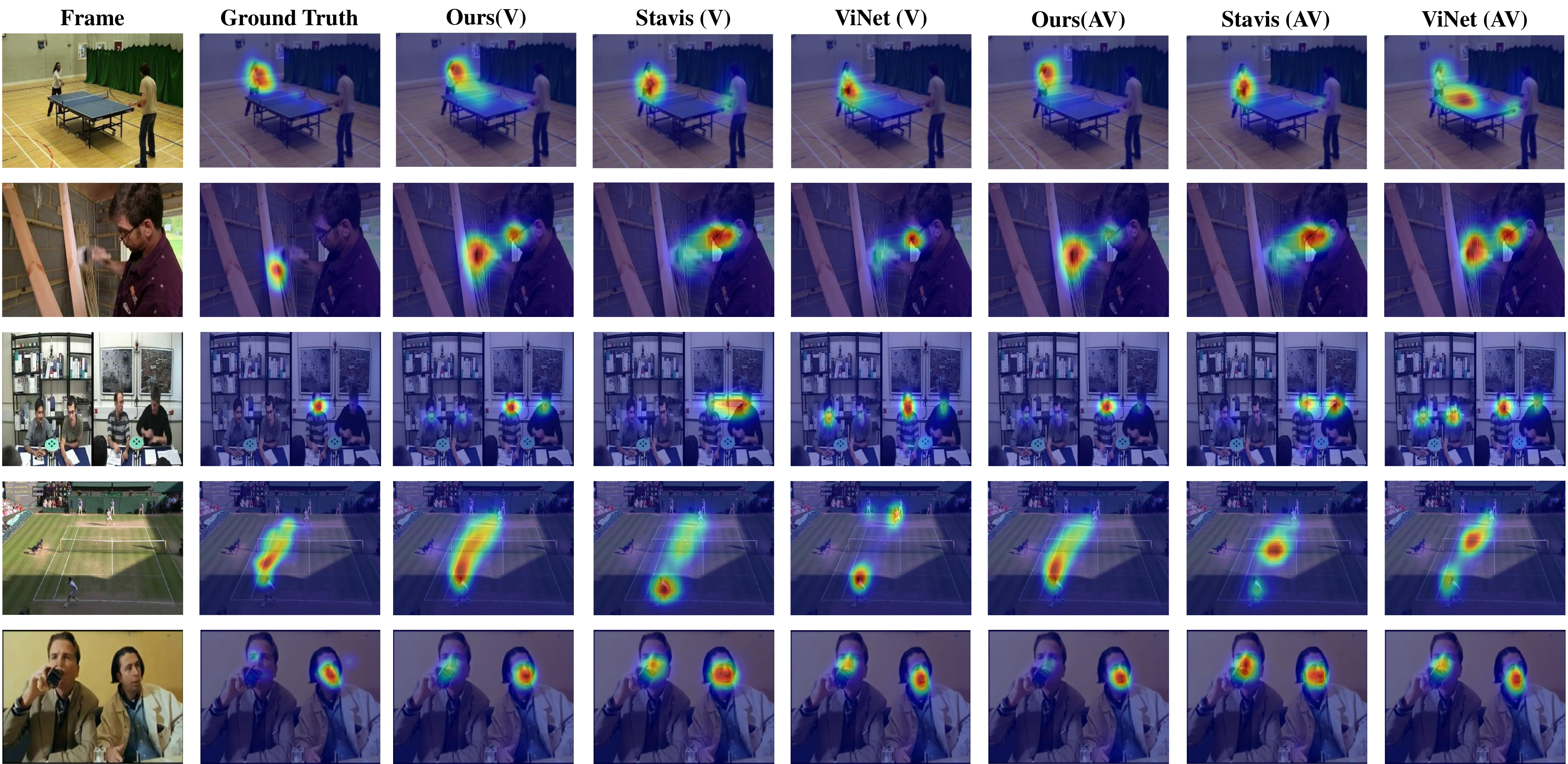}
\caption{Sample frame from Coutrot1, Coutrot2 and DIEM databases with their eye-tracking data, and the corresponding ground truth, AVRSP, and other state-of-the-art audio-visual saliency maps for comparisons.}
\label{fig:visualization}
\end{figure*}

In addition, we also provide the visual comparison results as shown in Figure \ref{fig:visualization}, where the top parts is the original frame sample and their ground truth eye-tracking data. The middle and bottom parts show the visualization results for visual-only and audio-visual version of all methods, and it is evident that our results are closer to the ground truth.

\subsection{Ablation Study}
Tables \ref{table:ablation study about multi-scale enhancement method}, \ref{table:ablation study about num of ms&mrg}, and \ref{table:ablation study about av fusion method} present the ablation studies under different configurations of AVRSP. All experiments are conducted by training on the DIEM and AVAD training sets and evaluating on their validation sets.

\begin{table*}[!t]
\renewcommand{\arraystretch}{1.1}
\caption{ABLATION STUDY OF MULTI-SCALE FEATURE ENHANCEMENT METHODS.}
\label{table:ablation study about multi-scale enhancement method}
\centering
\scalebox{1.2}{
\begin{tabular}{c|ccc|cccc|cccc}
\hline
\multirow{2}{*}{Model} & \multicolumn{3}{c|}{Configuration}   & \multicolumn{4}{c|}{DIEM}                                         & \multicolumn{4}{c}{AVAD}                                          \\
                       & Audio      & MS         & MRG        & SIM$\uparrow$  & CC$\uparrow$   & NSS$\uparrow$ & AUC-J$\uparrow$ & SIM$\uparrow$  & CC$\uparrow$   & NSS$\uparrow$ & AUC-J$\uparrow$ \\ \hline
\multirow{8}{*}{AVRSP} &            &            &            & 0.519          & 0.628          & 2.38          & 0.901           & 0.514          & 0.673          & 3.56          & 0.927           \\
                       &            & \checkmark &            & 0.533          & 0.654          & 2.66          & 0.904           & 0.529          & 0.687          & 3.74          & 0.932           \\
                       &            &            & \checkmark & 0.538          & 0.647          & 2.53          & 0.902           & 0.526          & 0.691          & 3.69          & 0.930           \\
                       &            & \checkmark & \checkmark & \textbf{0.545} & \textbf{0.668} & \textbf{2.67} & \textbf{0.907}  & \textbf{0.533} & \textbf{0.697} & \textbf{3.75} & \textbf{0.933}  \\ \cline{2-12} 
                       & \checkmark &            &            & 0.539          & 0.663          & 2.57          & 0.905           & 0.528          & 0.683          & 3.69          & 0.922           \\
                       & \checkmark & \checkmark &            & 0.551          & 0.676          & 2.72          & 0.908           & 0.540          & 0.696          & 3.77          & 0.932           \\
                       & \checkmark &            & \checkmark & 0.546          & 0.673          & 2.73          & 0.907           & 0.544          & 0.692          & 3.72          & 0.930           \\
                       & \checkmark & \checkmark & \checkmark & \textbf{0.559} & \textbf{0.685} & \textbf{2.78} & \textbf{0.911}  & \textbf{0.549} & \textbf{0.704} & \textbf{3.84} & \textbf{0.933}  \\ \hline
\end{tabular}
}
\end{table*}

\begin{table*}[!t]
\renewcommand{\arraystretch}{1.1}
\caption{ABLATION STUDY OF DIFFERENT NUMBERS OF MS\&MRG PAIRS}
\label{table:ablation study about num of ms&mrg}
\centering
\scalebox{1.1}{
\begin{tabular}{c|c|c|c|cccc}
\hline
                               & Number & Input                     & Output                     & CC$\uparrow$   & NSS$\uparrow$ & AUC-J$\uparrow$& SIM$\uparrow$  \\ \hline
\multirow{5}{*}{Visual-only}  & 4              & $X^{3},X^{2},X^{1},X^{0}$ & $X^{3},X^{2},X^{1},X^{0}$  & 0.665          & \textbf{2.68} & 0.905          & 0.540          \\
                               & 3              & $X^{3},X^{2},X^{1},X^{0}$ & $X^{3},X^{2},X^{1}$        & 0.668          & 2.67          & \textbf{0.907} & \textbf{0.545} \\
                               & 2              & $X^{3},X^{2},X^{1}$       & $X^{3},X^{2}$              & \textbf{0.669} & 2.64          & 0.904          & 0.541          \\
                               & 1              & $X^{3},X^{2}$             & $X^{3}$                    & 0.661          & 2.60          & 0.904          & 0.536          \\
                               & 0              & $Null$                      & $Null$                       & 0.652          & 2.54          & 0.900          & 0.528          \\ \hline
\multirow{5}{*}{Audio-Visual}  & 4            & $X^{3},X^{2},X^{1},X^{0}$ & $X^{3},X^{2},X^{1},X^{0}$    & 0.683          & 2.75          & 0.909          & 0.556          \\
                               & 3            & $X^{3},X^{2},X^{1},X^{0}$ & $X^{3},X^{2},X^{1}$          & \textbf{0.685} & 2.78          & \textbf{0.911} & \textbf{0.559} \\
                               & 2            & $X^{3},X^{2},X^{1}$       & $X^{3},X^{2}$                & 0.681          & \textbf{2.80} & 0.907          & 0.554          \\
                               & 1            & $X^{3},X^{2}$             & $X^{3}$                      & 0.676          & 2.73          & 0.908          & 0.554          \\
                               & 0            & $Null$                      & $Null$                        & 0.670          & 2.64          & 0.903          & 0.547          \\ \hline
\end{tabular}
}

\end{table*}

\textbf{Effectiveness of the MS and MRG Modules.} In the AVRSP model, the MS (Multi-scale feature Synergy) and MRG (Multi-scale Regulator Gate) modules play a significant role in enhancing the model's performance. The results from the ablation study presented in Table \ref{table:ablation study about multi-scale enhancement method} provide a macro-level understanding of the effectiveness of these two modules.

The table shows that as the MS and MRG modules are progressively added, there is a noticeable improvement in the performance metrics of the AVRSP model. This indicates that these modules are crucial to improving the accuracy and robustness of the video saliency prediction model. The MS module, by integrating visual features from different encoding stages, enhances the model's ability to represent objects at various scales. This module effectively improves saliency prediction, allowing the model to capture and predict salient regions more accurately in videos.

Simultaneously, the MRG module optimizes the utilization of multi-scale visual features by channeling essential fusion information. This allows the model to better integrate audio and visual information, thereby increasing the precision and consistency of the saliency predictions. When both MS and MRG modules are employed together, the model achieves the best performance, demonstrating the synergistic effect of these modules. This synergy enables the model to excel in handling complex dynamic and diverse content in audio-visual datasets.

The results from Table \ref{table:ablation study about multi-scale enhancement method} highlight the critical role of the MS and MRG modules in the AVRSP model. By enhancing the representation of multi-scale features and optimizing the fusion of audio-visual information, these modules significantly improve the overall performance of the model, making it more effective in video saliency prediction tasks. This improvement is particularly important for handling audio-visual datasets with complex dynamics and diverse content.

\begin{table*}[!t]
\renewcommand{\arraystretch}{1.1}
\caption{ABLATION STUDY OF DIFFERENT MULTI-MODAL FEATURE FUSION METHODS}
\label{table:ablation study about av fusion method}
\centering
\scalebox{1.2}{
\begin{tabular}{l|cccc|cccc}
\hline
\multirow{2}{*}{Fusion Methods} & \multicolumn{4}{c|}{DIEM}                                         & \multicolumn{4}{c}{AVAD}                                          \\
                                & SIM$\uparrow$  & CC$\uparrow$   & NSS$\uparrow$ & AUC-J$\uparrow$ & SIM$\uparrow$  & CC$\uparrow$   & NSS$\uparrow$ & AUC-J$\uparrow$ \\ \hline
Visual-only baseline            & 0.546          & 0.668          & 2.67          & 0.907           & 0.533          & 0.697          & 3.75          & 0.933           \\ \hline
Element-wise addition           & 0.547          & 0.665          & 2.54          & 0.897           & 0.531          & 0.690          & 3.61          & 0.920           \\
Element-wise multiplication     & 0.542          & 0.658          & 2.53          & 0.903           & 0.528          & 0.688          & 3.61          & 0.913           \\
Concatenation                   & 0.545          & 0.663          & 2.60          & 0.895           & 0.530          & 0.688          & 3.69          & 0.915           \\ \hline
Bilinear                        & 0.553          & 0.672          & 2.66          & 0.906           & 0.537          & 0.696          & 3.74          & 0.929           \\
MBT\cite{nagrani2021attention}                             & 0.547          & 0.673          & 2.73          & 0.908           & 0.541          & 0.697          & 3.78          & 0.927           \\
RAVF(Ours)                      & \textbf{0.559} & \textbf{0.685} & \textbf{2.78} & \textbf{0.911}  & \textbf{0.549} & \textbf{0.704} & \textbf{3.84} & \textbf{0.933}  \\ \hline
\end{tabular}
}
\end{table*}

\begin{figure}[ht]
\centering
\includegraphics[width=3.5in]{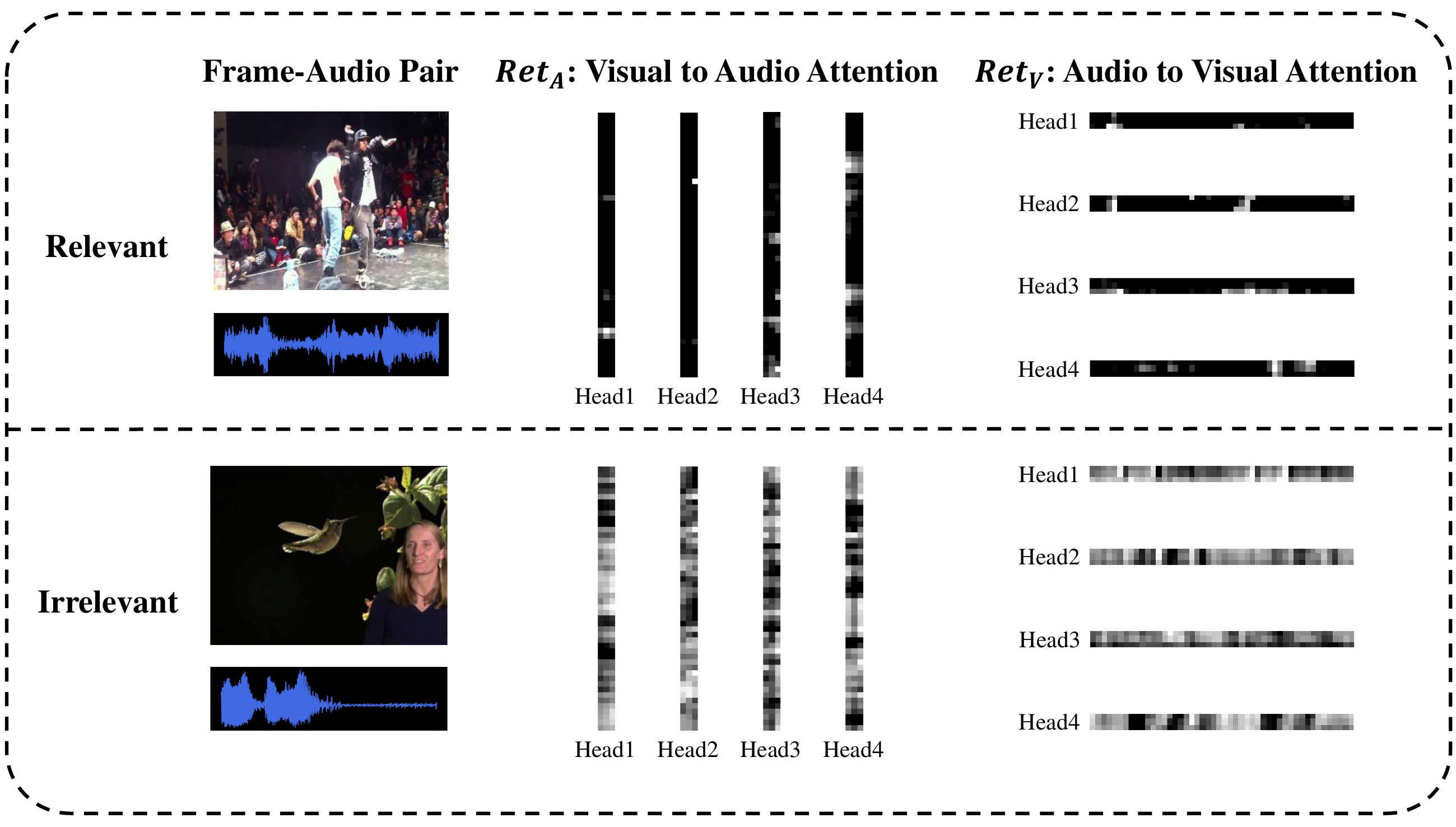}
\caption{The visualization attention map of $Ret_A$ and $Ret_V$ in both relevant and irrelevant configurations. The brighter the area, the closer the value is to 1, indicating greater relevance. Conversely, darker areas indicate less relevance.}
\label{fig:retention}
\end{figure}

\textbf{Evaluation of the Audio-Visual Fusion Method.} To evaluate the performance of various feature fusion approaches in our proposed model, a baseline configuration "Visual-only" was established, which relies solely on the visual backbone, Multi-scale (MS), Multi-scale Regulator Gate (MRG), and a Saliency Decoder, as detailed in Table \ref{table:ablation study about av fusion method}. We explored several fusion techniques, including simple element-wise operations and more sophisticated learning-based methods. Among the simple fusion methods, element-wise addition, element-wise multiplication, and concatenation, even none of them significantly surpassed the performance of the "visual-only" baseline. In the learning-based category, the Bilinear and MBT\cite{nagrani2021attention} showed some improvements but still did not reach the effectiveness of our specially designed Relevance-guided Audio-Visual feature Fusion (RAVF). The RAVF method excelled over all other tested methods, demonstrating a superior ability to effectively integrate audio and visual cues. For example, as shown in Figure\ref{fig:av relevance example}, during scenes with background music, our RAVF-based model minimizes the influence of irrelevant audio and focuses more sharply on crucial visual elements, capturing essential parts of the video accurately. In segments with a narrative, such as a woman speaking, it prioritizes the speaker's presence, emphasizing relevant visual details closely associated with the audio. In scenarios where only ambient sounds like bird chirping are present, the model adeptly shifts its focus towards relevant visual contexts.

In the visualization of the attention maps $Ret_A$ and $Ret_V$ for both relevant and irrelevant configurations, as shown in Figure \ref{fig:retention}, there are significant differences in RetA (Visual to Audio Attention) and RetV (Audio to Visual Attention). When audio and video content is related, these attention maps exhibit values close to 1 (brighter), indicating a strong correlation between audio and visual information. This demonstrates the module's ability to effectively identify and emphasize relevant multimodal features. Conversely, for irrelevant audio-visual pairs, the values of RetA and RetV are close to 0 (darker), showcasing the system's capacity to ignore irrelevant audio information and prevent it from interfering with the processing of visual content. These results affirm the module's capability to accurately differentiate between relevant and irrelevant information during audio-visual feature fusion, underscoring its effectiveness in multimodal integration.

\section{Conclusion}
In this study, we introduce AVRSP, a novel multimodal video saliency prediction network designed to address the challenges associated with the incongruity between visual and auditory modalities. To mitigate the issue of visual-audio mismatch, we devised a strategic approach for the fusion of visual and audio features, underpinned by the correlation between these two modalities. Furthermore, to enhance the model's predictive capability across objects of varying sizes, we have developed a multi-scale feature synergy module. Rigorous evaluations were performed using both vision-centric and audio-visual datasets that are widely acknowledged within the research community. The experimental results show that the AVRSP model achieves better performance improvement compared with the existing methods.

\bibliographystyle{IEEEtran}
\bibliography{reference.bib}
\newpage
\end{document}